\documentclass[letterpaper]{article} 
\usepackage{aaai25}  
\usepackage{times}  
\usepackage{helvet}  
\usepackage{courier}  
\usepackage[hyphens]{url}  
\usepackage{graphicx} 
\usepackage{natbib}  
\usepackage{caption} 
\usepackage{algorithm}
\usepackage{algorithmic}

\usepackage[utf8]{inputenc} 
\usepackage[T1]{fontenc}    
\usepackage{url}            
\usepackage{booktabs}       
\usepackage{amsfonts}       
\usepackage{nicefrac}       
\usepackage{microtype}      
\usepackage{xcolor}         
\usepackage{xpatch}
\usepackage{graphicx} 
\usepackage{amsmath} 
\usepackage{color} 
\usepackage{array} 
\usepackage{amssymb} 
\usepackage{pifont} 
\usepackage{threeparttable} 
\usepackage{caption} 
\usepackage{makecell} 
\usepackage{multirow} 
\usepackage{enumitem} 
\usepackage{xpatch}

\usepackage{newfloat}
\usepackage{listings}
\DeclareCaptionStyle{ruled}{labelfont=normalfont,labelsep=colon,strut=off} 
\floatstyle{ruled}
\newfloat{listing}{tb}{lst}{}
\floatname{listing}{Listing}
\title{\Huge Atmos-Bench: \\3D Atmospheric Structures for Climate Insight}
\author {
    Tianchi Xu\textsuperscript{\rm 1}
}
\affiliations {
    \textsuperscript{\rm 1}Institute of Remote Sensing and Geographical Information Systems, Peking University\\
    tianchixu@stu.pku.edu.cn
}

\usepackage{bibentry}

\begin{document}

\maketitle
\maketitle

\begin{abstract}
  Atmospheric structure, represented by backscatter coefficients (BC) recovered from satellite LiDAR attenuated backscatter (ATB), provides a volumetric view of clouds, aerosols, and molecules, playing a critical role in human activities, climate understanding, and extreme weather forecasting.
  Existing methods often rely on auxiliary inputs and simplified physics-based approximations, and lack a standardized 3D benchmark for fair evaluation.
  However, such approaches may introduce additional uncertainties and insufficiently capture realistic radiative transfer and atmospheric scattering–absorption effects. 
  To bridge these gaps, we present Atmos-Bench: the first 3D atmospheric benchmark, along with a novel FourCastX: Frequency‑enhanced Spatio‑Temporal MiXture‑of‑Experts Network.
  that
  a) generates 921,600 image slices from 3D scattering volumes simulated at 532 nm and 355 nm, by coupling WRF with an enhanced COSP simulator over 384 land–ocean time steps, yielding high-quality voxel-wise references;
  b) embeds ATB–BC physical constraints into model architecture, promoting energy consistency during restoration; 
  c) achieves consistent improvements on the Atmos-Bench dataset across both 355\,nm and 532\,nm bands, outperforming state-of-the-art baseline models without relying on auxiliary inputs.
  Atmos-Bench establishes a new standard for satellite-based 3D atmospheric structure recovery and paves the way for deeper climate insight.
\end{abstract}

%

\section{Introduction}

Understanding Earth's climate system requires high-fidelity 3D characterization of clouds, aerosols, and gas molecules, which influence the planet’s radiative balance through scattering and absorption \cite{10.3389/frsen.2024.1477503}. These processes drive weather systems, shape long-term climate trends, and intensify extreme events such as heatwaves, floods, and storms—many of which are increasingly linked to human activities\cite{yin2024precipitationnowcastingusingphysics}. Remote sensing plays a central role in monitoring these dynamics\cite{NEURIPS2023_11822e84}. Active satellite sensors, such as spaceborne LiDARs (e.g., CALIPSO CALIOP, EarthCARE ATLID)\cite{amt-17-5301-2024}, provide vertically resolved measurements of attenuated backscatter (ATB), from which backscatter coefficients (BC) can be estimated\cite{2023Retrieving}. BC serves as a proxy for the atmospheric structure. However, due to signal attenuation, especially through thick clouds, the raw ATB often suffers from missing or degraded information\cite{rs16152735}\cite{rs17071215}.

Recent advances in deep learning and computer vision have enabled effective solutions to inverse problems such as image inpainting, denoising, and volumetric reconstruction\cite{Zhang_2018_CVPR}. Architectures including convolutional neural networks (CNNs), generative adversarial networks (GANs), and transformers have demonstrated strong generalization capabilities by learning spatial priors from large-scale visual datasets\cite{7115171}. In the remote sensing domain, foundation models such as SatMAE, SpectralGPT, RemoteCLIP, FlexiMo, and SkySense highlight the potential of self-supervised pretraining on satellite imagery\cite{cong2023satmaepretrainingtransformerstemporal}\cite{10.1109/TPAMI.2024.3362475}\cite{liu2024remoteclipvisionlanguagefoundation}\cite{2023SkySense}. However, these models are typically not trained on atmospheric datasets and are not explicitly designed for recovering the 3D structure of the atmosphere. Many architectures prioritize visual fidelity, which may limit their ability to capture physically consistent atmospheric features. Moreover, domain-specific characteristics, such as signal attenuation, vertical continuity, and spatial sparsity, are not always fully reflected in model design.

As a result, while current methods have achieved impressive performance in general visual tasks, their applicability to atmospheric remote sensing remains an open challenge. The unique nature of volumetric atmospheric data, particularly from active satellite sensors, introduces complexities that are not fully addressed by existing architectures.This underscores the need for models that are explicitly optimized for atmospheric reconstruction, incorporating domain-informed design principles.

In this work, we introduce \textbf{FourCastX}, a unified framework that integrates deep generative modeling with physics-informed learning~\cite{wang2020physicsinformedneuralnetworksuper} for recovering backscatter coefficients from ATB profiles. FourCastX employs a frequency-enhanced encoder–decoder backbone based on Fast Fourier Convolution (FFC)~\cite{Chi_2020_FFC} to extract multi-scale spatial representations, incorporates a VisionLSTM module for temporal and contextual memory encoding, and utilizes a spatial cross-attention decoder for precise reconstruction. The model is trained with a composite objective that includes reconstruction loss for data fidelity, a differentiable physics constraint enforcing the ATB–BC energy relationship, and perceptual regularization—achieving strong performance without auxiliary inputs.

To enable robust and reproducible evaluation, we present \textbf{Atmos-Bench}, the first large-scale benchmark dataset for 3D atmospheric backscatter recovery~\cite{essd-14-3821-2022}, along with \textbf{FourCastX}, a novel physics-informed, residual-aware, frequency-enhanced architecture. Atmos-Bench is constructed from high-resolution WRF simulations coupled with an enhanced COSP forward model, generating 384 volumetric scenes at two wavelengths (355\,nm and 532\,nm). From these, we extract over 460{,}000 voxel-aligned ATB–BC slice pairs across diverse land, oceanic, and atmospheric conditions. 
We evaluate FourCastX on Atmos-Bench and demonstrate that it consistently outperforms traditional radiative transfer inversions and data-driven baselines, yielding physically consistent and spatially coherent reconstructions. In
summary, this work:
\begin{itemize}
  \item Establishes \textbf{Atmos-Bench}, a standardized 3D \textbf{Atmos}pheric structure recovery \textbf{Bench}mark;
  \item Defines a unified evaluation protocol for systematic and reproducible comparison;
  \item Introduces \textbf{FourCastX}, a state-of-the-art generative baseline with embedded physical constraints;
  \item Facilitates climate understanding via high-fidelity atmospheric reconstructions.
\end{itemize}

The rest of the paper is organized as follows: In Sec.~\ref{sec:related}, we review existing physics-based and learning-based approaches for atmospheric volumetric recovery. Sec.~\ref{sec:dataset} introduces the construction and key characteristics of the Atmos-Bench dataset, including its simulation sources and scene diversity. In Sec.~\ref{sec:model}, we detail the architecture of our proposed FourCastX framework and the design of its loss functions, including physics-informed constraints. Sec.~\ref{sec:experiments} presents comprehensive experiments on standardized dataset, including comparisons with state-of-the-art baselines. Finally, in Sec.~\ref{Conclusion}, we discuss the limitations of our approach and outline promising directions for future research.

\section{Related Works}
\label{sec:related}

\begin{figure}[!htbp]
  \centering
  \includegraphics[width=1\linewidth]{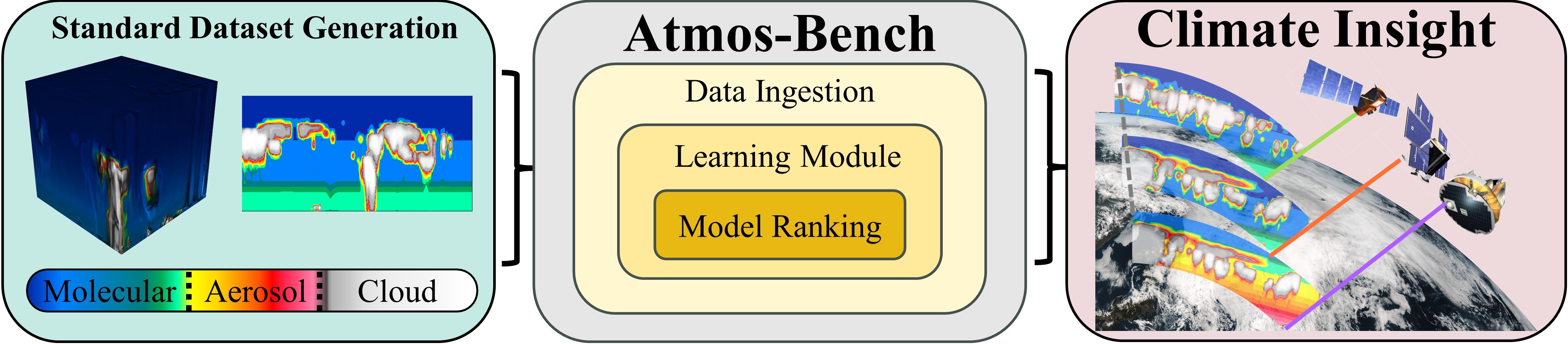}
  \caption{Overview of Atmos-Bench: generation of voxel-aligned 3D scenes via WRF–COSP simulations at two wavelengths (left), extraction of ATB–BC slice pairs and unified evaluation pipeline (center), and physics-informed restoration enabling downstream climate insights (right).}
  \label{fig:Atmos-Bench}
\end{figure}

Understanding the vertical structure of atmospheric components—such as molecular scattering, aerosols, and clouds—is crucial for climate modeling, satellite retrieval, and weather forecasting. However, standardized datasets and unified evaluation pipelines for benchmarking learning-based methods in this domain are still lacking. To address this gap, we propose \textbf{Atmos-Bench}, a comprehensive benchmark designed to facilitate fair comparison, promote reproducibility, and accelerate research at the intersection of atmospheric science and machine learning. An overview of the proposed Atmos-Bench framework is illustrated in Figure~\ref{fig:Atmos-Bench}.

\paragraph{Remote Sensing Benchmarks and Foundation Models} Figure~\ref{fig:Atmos-Bench} illustrates the end-to-end Atmos-Bench workflow, and recent efforts have introduced standardized benchmarks to facilitate the development and evaluation of remote sensing foundation models. GEO-Bench provides a suite of classification and segmentation tasks tailored for Earth monitoring applications\cite{NEURIPS2023_a0644215}. FoMo-Bench focuses on forest monitoring, encompassing classification, segmentation, and object detection tasks\cite{bountos2025fomomultimodalmultiscalemultitask}. PhilEO Bench offers a framework for evaluating geospatial foundation models on tasks such as building density estimation and land cover classification. These benchmarks have been instrumental in advancing models like SpectralGPT, which leverages a 3D generative pretrained transformer architecture for spectral remote sensing data, and SkySense, a multi-modal model incorporating temporal sequences of optical and SAR data for comprehensive Earth observation\cite{10332123}\cite{2023SkySense}. Additionally, Panopticon introduces an any-sensor foundation model capable of processing arbitrary sensor inputs by treating images from different sensors as natural augmentations and employing cross-attention mechanisms\cite{waldmann2025panopticonadvancinganysensorfoundation}. While these initiatives have significantly contributed to the field, they primarily target 2D surface variables or classification/regression tasks, highlighting the need for benchmarks addressing full 3D atmospheric structure recovery.

\paragraph{Image Inpainting and Volumetric Restoration} Advances in deep learning have led to powerful models for image restoration tasks. Architectures such as Cross Aggregation Transformer (CAT), TSFormer, DiffIR, DDS2M, VmambaIR, MambaIRv2, UIR-LoRA, FrePrompter, EchoIR, and EnsIR have demonstrated remarkable capabilities in image inpainting and denoising\cite{zheng2022cross}\cite{miao2023dds2m}\cite{miao2023dds2mselfsuperviseddenoisingdiffusion}\cite{shi2024vmambairvisualstatespace}\cite{guo2025mambairv2attentivestatespace}\cite{guo2025mambairv2attentivestatespace}\cite{WU2025111223}\cite{he2024echoiradvancingimagerestoration}\cite{sun2024ensirensemblealgorithmimage}\cite{suvorov2022resolution}. These models excel in capturing spatial dependencies and restoring missing or corrupted image regions. However, their application to volumetric reconstruction, particularly in the context of atmospheric data, remains limited. The unique challenges posed by atmospheric observations—such as sparsity, noise, and the necessity to preserve physical constraints like energy conservation—necessitate models that can effectively handle 3D structures and integrate domain-specific knowledge.

\paragraph{Remote Sensing Foundation Datasets} The development of foundation models in remote sensing has been propelled by large-scale datasets. SpectralGPT was trained on one million spectral remote sensing images, enabling it to handle varying sizes, resolutions, and time series data\cite{10490262}. SkySense utilized a curated multi-modal dataset with 21.5 million temporal sequences, incorporating both optical and SAR data to enhance its generalization capabilities\cite{guo2024skysensemultimodalremotesensing}. Prithvi-EO-2.0 leveraged 4.2 million global time series samples from NASA's Harmonized Landsat and Sentinel-2 data archive, incorporating temporal and location embeddings for improved performance across geospatial tasks\cite{szwarcman2025prithvieo20versatilemultitemporalfoundation}. Despite the richness of these datasets, there remains a gap in datasets specifically designed for 3D atmospheric backscatter recovery, underscoring the importance of developing benchmarks like Atmos-Bench that cater to the unique demands of volumetric atmospheric retrieval\cite{10.3389/frsen.2024.1477503}.

\section{Atmos-Bench Dataset Generation}
\label{sec:dataset}

\begin{figure*}[!htbp]
  \centering
  \includegraphics[width=0.95\linewidth]{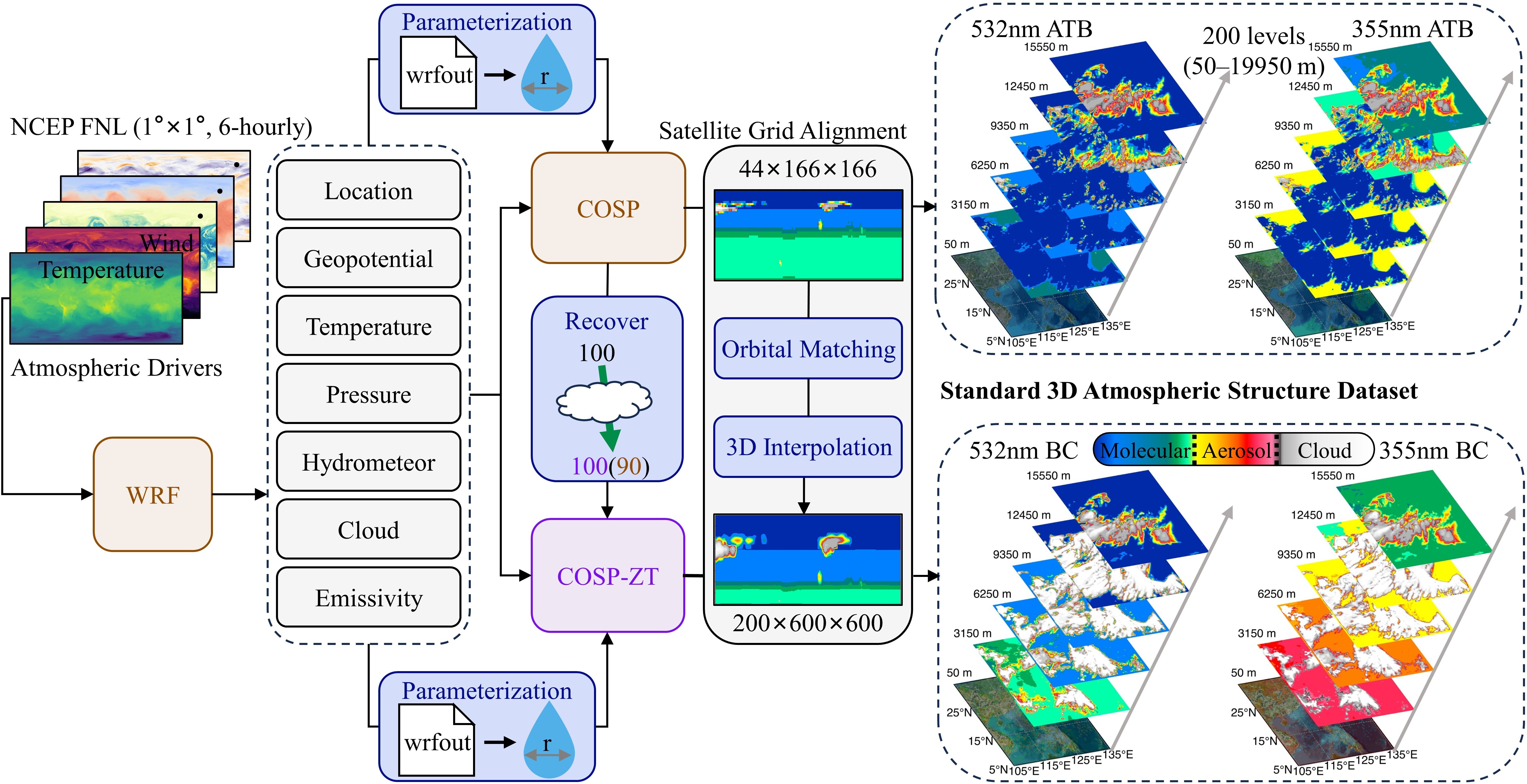}
  \caption{Atmos-Bench dataset generation pipeline. Starting from NCEP FNL reanalysis (1°, 6-hourly, 33 levels), WRF simulates regional atmospheric states (18 km, 44 levels). Outputs are passed through standard and enhanced COSP simulators to produce ATB and BC backscatter. These are aligned to a 5 km satellite grid (200 levels) and sliced zonally to form paired 2D ATB/BC images.}
  \label{fig:dataset_pipeline}
\end{figure*}

We develop \textbf{Atmos-Bench} by integrating state-of-the-art atmospheric reanalysis data, high-resolution numerical weather simulations, and advanced satellite forward modeling into a unified simulation pipeline~\cite{10.3389/frsen.2024.1477503} (Fig.~\ref{fig:dataset_pipeline}). Starting from global reanalysis inputs (e.g., ERA5), we use the WRF model to produce high-resolution regional atmospheric states, including 3D temperature, pressure, humidity, and hydrometeor fields. These outputs are then processed by an enhanced version of the COSP simulator, where we extend the lidar module to generate not only standard attenuated total backscatter (ATB) and intrinsic backscatter coefficient (BC), but also an additional non-absorbing (enhanced) backscatter product that removes the effects of molecular and aerosol attenuation. All simulated signals are subsequently projected onto real satellite orbital grids and segmented into 2D zonal image patches to support deep learning applications \cite{ he2024residual, he2024generalized, he2024mutual, xu2023spatio, he2024uncertainty, he2025co, he2024epl}.

The final product of this pipeline is the \textbf{Standard 3D Atmospheric Structure Dataset}, which contains two wavelength-specific subsets: 532 nm and 355 nm. Each subset includes paired ATB and BC fields, enabling learning under both physically attenuated and idealized non-attenuated conditions.

\subsection{Reanalysis Inputs and WRF Configuration}
We initialize our pipeline with the NCEP FNL reanalysis dataset, which provides global atmospheric variables at 1° horizontal resolution, 6-hourly intervals, and 33 vertical levels up to 20 km altitude\cite{atmos12040481}. The key drivers ingested include temperature, geopotential height, zonal and meridional wind components, pressure, humidity, cloud fraction, and surface emissivity. 
These are input into the Weather Research and Forecasting (WRF) model over a regional domain spanning 104.5°E–135.5°E and 4.9°N–34.9°N. Two simulation periods (2020‑07‑02 to 2020‑07‑12 and 2020‑08‑09 to 2020‑08‑13), each with a preceding 5‑day spinup (excluded from analysis) to ensure simulation quality, yield a total of 16 days (384 hourly snapshots) on an 18 km grid with 44 vertical levels. The WRF configuration employs Morrison double-moment microphysics\cite{TheResponseTimeoftheTemperatureoftheEquatorialTropospheretoENSOHeating}, Dudhia shortwave, RRTM longwave radiation schemes, Kain–Fritsch cumulus parameterization, and the YSU planetary boundary layer scheme\cite{TIAN2017141}. The model outputs latitude/longitude grids, geopotential, temperature, pressure, hydrometeor mixing ratios and number concentrations (for effective radius calculation), cloud fraction, and emissivity.

\subsection{COSP and Zero-Tau Forward Simulation}
WRF outputs are processed by two configurations of the COSP (Cloud Feedback Model Intercomparison Project Observation Simulator Package) simulator at 532 nm and 355 nm wavelengths\cite{gmd-11-77-2018}. The standard configuration generates ATB by simulating realistic satellite-observed signals with full atmospheric attenuation. We develop a zero-optical-depth configuration, COSP-ZT, which disables attenuation by setting extinction to zero in the forward model\cite{gmd-16-1359-2023}. This enhancement yields intrinsic BC signals as clear proxies of atmospheric structure, complementing the radiatively attenuated ATB views. Both ATB and BC are generated at the native WRF resolution before further alignment.

\subsection{Grid Alignment and Image Slice Extraction}
To support the pivotal and ongoing EarthCARE mission, we align the ATB and BC volumes to its observation-centric grid through orbital matching and three-dimensional interpolation\cite{amt-17-5301-2024}, producing data at 5 km horizontal resolution with 200 vertical levels up to 20 km altitude. From these aligned volumes, we extract meridional slices to generate 921,600 paired images (200×600 pixels) across two wavelengths and two signal types. This large-scale, physically grounded dataset serves as a benchmark for training and fairly evaluating generative reconstruction models and physics-aware inversion methods.

\section{Model}
\label{sec:model}

\begin{figure*}[ht]
    \centering
    \includegraphics[width=1\linewidth]{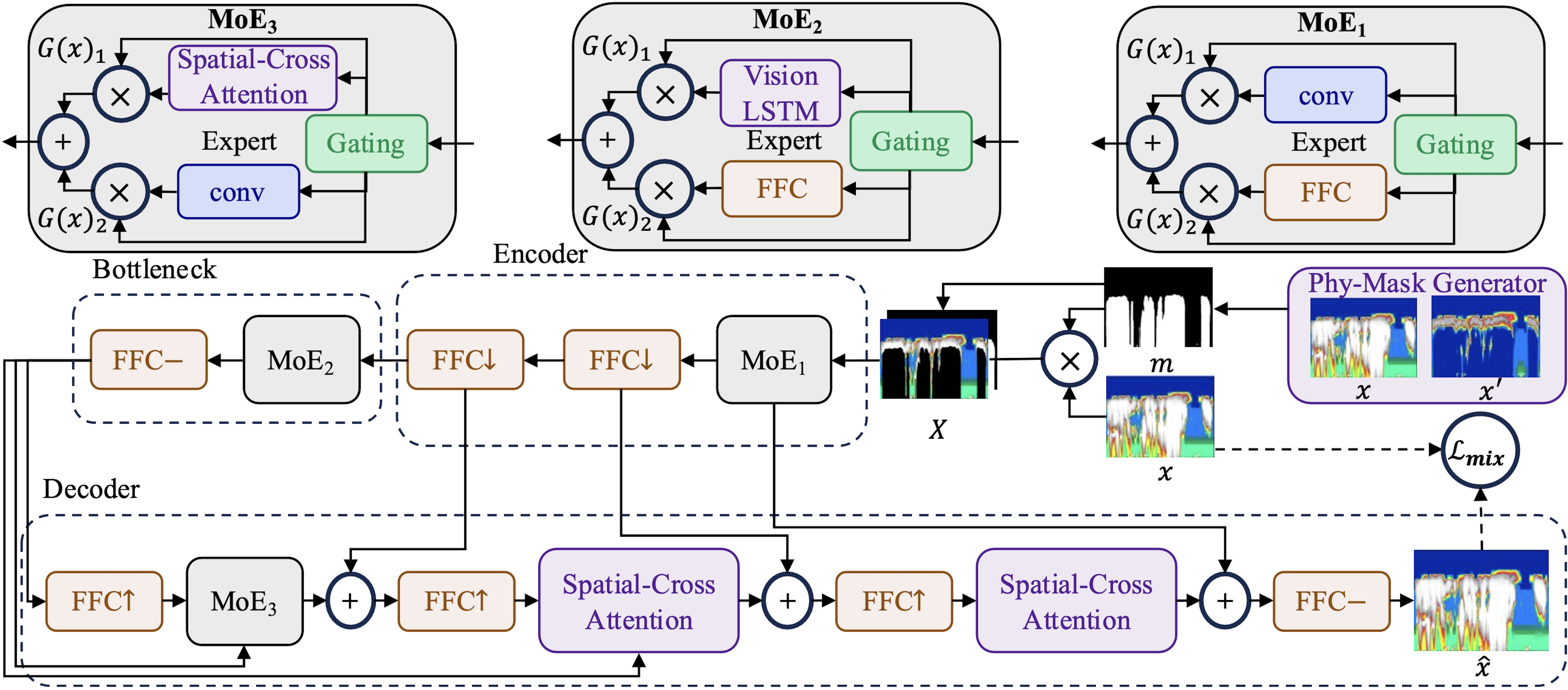}
    \caption{Architecture of FourCastX. The encoder–decoder backbone employs FFC modules for frequency-enhanced representation. Gated Mixture-of-Experts (MoE\textsubscript{1-3}) dynamically select among experts: convolution, FFC, VisionLSTM, and Spatial-Cross Attention. Physics-informed mask generation facilitates hybrid supervision via \(\mathcal{L}_{\text{mix}}\).}
    \label{fig:model}
\end{figure*}

We propose \textbf{FourCastX}, a physics-aware generative framework designed to restore the intrinsic backscattering coefficient (BC) from attenuated ATB observations. As illustrated in Figure~\ref{fig:model}, our model adopts a U-shaped encoder–decoder structure enhanced by a modular Mixture-of-Experts (MoE) mechanism and frequency-aware processing modules. The key components are summarized as follows:

\begin{itemize}[left=0pt]
  \item \textbf{Multi-scale Encoder–Decoder with FFC:} We adopt Fast Fourier Convolution (FFC) blocks~\cite{Chi_2020_FFC} throughout the encoder, bottleneck, and decoder to capture both local texture and long-range context. The encoder down-samples the input with \texttt{FFC$\downarrow$}, while the decoder upsamples with \texttt{FFC$\uparrow$}.
  
  \item \textbf{Mixture-of-Experts (MoE) Modules:} Each stage of the encoder and decoder incorporates gated MoE blocks that dynamically combine different processing pathways. Specifically:
    \begin{itemize}
      \item \textbf{MoE\textsubscript{1}} mixes FFC and standard convolution.
      \item \textbf{MoE\textsubscript{2}} leverages FFC and VisionLSTM~\cite{alkin2024vision} to model spatiotemporal dependencies.
      \item \textbf{MoE\textsubscript{3}} fuses convolution and Spatial-Cross Attention~\cite{WOS:000971500203043} to preserve spatial structure in reconstruction.
    \end{itemize}

  \item \textbf{Spatial-Cross Attention in Decoder:} Global features extracted from the bottleneck are fused with decoder layers via Spatial-Cross Attention to guide high-resolution recovery with contextual awareness.

  \item \textbf{Physics-informed Supervision:} A physics-based mask generator computes masked regions to guide learning. A hybrid image \(\hat{x}\) is created from mixing physical and predicted content using the generated mask \(m\), and the model is supervised with a physics-mixing loss \(\mathcal{L}_{\text{mix}}\).

\end{itemize}

The training objective combines evidential regression, adversarial, perceptual~\cite{WOS:000861612700053}, feature-matching, and \(L_1\) losses~\cite{WOS:000389383900043} to enforce physical consistency and reconstruction fidelity.

\subsection{Multi-Path Fourier-aware Encoder}

The Multi-Path Fourier-aware Encoder processes a two-channel input to produce a hierarchical, multi-scale feature representation, where \(H\) and \(W\) denote the height and width of the input image.
\begin{equation}
x = (\text{masked\_image}, \text{mask}) \in \mathbb{R}^{2 \times H \times W}
\label{eq:x_input}
\end{equation}
Initially, a \(3 \times 3\) convolutional layer projects the input into 64 feature channels, producing the initial feature map \(e_0\):
\begin{equation}
e_0 = \mathrm{Conv2d}_{2 \to 64}(x)
\label{eq:e0}
\end{equation}
where \(e_0 \in \mathbb{R}^{B \times 64 \times H \times W}\). The encoder then applies three successive processing blocks indexed by \(i = 1, 2, 3\), each of which consumes the downsampled output \(e_{i-1}^\downarrow\) from the previous layer. At \(i=1\), we use a Mixture-of-Experts Fast Fourier Convolution (MoE-FFC) block\cite{shazeer2017outrageouslylargeneuralnetworks}, while at \(i=2, 3\), standard Fast Fourier Convolution (FFC) blocks are applied:
\begin{equation}
e_i = 
\begin{cases}
\mathrm{MoE\text{-}FFC}_{C_i \to C_{i+1}}(e_{i-1}^\downarrow) & i=1 \\
\mathrm{FFC}_{C_i \to C_{i+1}}(e_{i-1}^\downarrow) & i=2,3
\end{cases}
\label{eq:ei}
\end{equation}
where \(e_i \in \mathbb{R}^{B \times C_{i+1} \times H_i \times W_i}\), and \(C_i, C_{i+1}\) denote the input and output channel widths, while \(H_i, W_i\) are the spatial dimensions after downsampling. The downsampled version of each feature map, denoted by \(e_i^\downarrow\), is computed via a strided convolution:
\begin{equation}
e_i^\downarrow = \mathrm{Conv2d}_{C_{i+1} \to C_{i+1}}^{\downarrow 2}(e_i) \quad i=0,1,2
\label{eq:downsample}
\end{equation}
where \(e_i^\downarrow \in \mathbb{R}^{B \times C_{i+1} \times \frac{H_i}{2} \times \frac{W_i}{2}}\). Throughout the encoder, the channel configuration is fixed as \((C_1, C_2, C_3, C_4) = (64, 128, 256, 512)\), defining the number of feature channels at each stage.

The MoE-FFC block dynamically routes features through three parallel experts: a standard convolutional expert, a frequency-aware FFC expert, and a ConvLSTM-based temporal expert\cite{WOS:000450913102103}. A soft gating mechanism adaptively selects and combines expert outputs based on the input content, allowing the encoder to emphasize spatial, spectral, or temporal cues depending on the scene. The FFC blocks decompose the features into global (Fourier) and local components, enabling efficient modeling of both long-range and fine-grained patterns. Through repeated applications of these blocks and strided downsampling, the encoder progressively reduces the spatial resolution while preserving rich semantic representations, forming a multi-scale hierarchical encoding structure. These representations are propagated to both the bottleneck and decoder, enabling efficient feature reuse via skip connections.

\subsection{Bottleneck Mixture of Experts}

At the lowest spatial resolution, which is \(1/8\) of the input size, the encoder produces a bottleneck feature map \(e\), as shown in Equation~\ref{eq:bottleneck_input}. This feature map is subsequently processed by a Mixture-of-Experts (MoE) module designed to balance spatial–spectral reasoning with temporal memory integration\cite{shazeer2017outrageouslylargeneuralnetworks}. The MoE module comprises two parallel experts: VisionLSTM, which leverages ConvLSTM to capture temporal consistency across slices\cite{WOS:000450913102103}, and FFCBlock, which employs Fourier convolution to extract both global spectral features and local spatial details\cite{WOS:000861612700053}.
\begin{equation}
e = e_3^\downarrow \in \mathbb{R}^{B \times 512 \times \frac{H}{8} \times \frac{W}{8}}
\label{eq:bottleneck_input}
\end{equation}

To adaptively fuse the outputs from the two experts, a lightweight gating network is employed to determine their relative contributions~\cite{WOS:001275019903043}. Specifically, the bottleneck feature map \(e\) is first passed through a Conv–Pool–FullyConnected pipeline to produce the routing weights \(w \in \mathbb{R}^{B \times 2}\), as shown in Equation~\ref{eq:gating}. This process involves applying a \(1 \times 1\) convolution (\(\mathrm{Conv}_{1 \times 1}\)) to reduce the channel dimension from 512 to a hidden dimension \(G\), followed by global average pooling (\(\mathrm{GAP}\)) to summarize spatial information into a compact vector of size \(G\). The resulting vector is then fed into a fully connected layer to generate the logits for the two experts. A softmax function is applied to obtain normalized routing weights~\cite{jain2024mixturenestedexpertsadaptive}. During training, Gaussian noise \(\epsilon \sim \mathcal{N}(0, \sigma^2)\) may be added to encourage exploration and improve generalization~\cite{liu2022gatingdropoutcommunicationefficientregularization}:
\begin{equation}
w = \mathrm{softmax}\left( \mathrm{FC}\left( \mathrm{GAP}\left( \mathrm{Conv}_{1 \times 1}(e) \right) + \epsilon \right) \right)
\label{eq:gating}
\end{equation}

The final bottleneck representation \(b \in \mathbb{R}^{B \times 512 \times \frac{H}{8} \times \frac{W}{8}}\) is computed as the weighted sum of the two expert outputs in Equation~\ref{eq:moe_output}, where \(E_1 = \text{VisionLSTM}\) and \(E_2 = \text{FFCBlock}\):
\begin{equation}
b = \sum_{i=1}^{2} w_i E_i(e)
\label{eq:moe_output}
\end{equation}
This adaptive fusion allows the network to balance temporal and spectral–spatial features based on each input slice, enhancing generalization\cite{han2025fusemoemixtureofexpertstransformersfleximodal}. The fused output \(b\) is then forwarded to the decoder. This MoE bottleneck design dynamically selects the most relevant feature processing pathway according to the input content at each spatial location\cite{WOS:001213927509018}.

\subsection{Spatial Cross-Attention Decoder}

The decoder reconstructs spatial resolution in three stages. At each stage \(i\), the previous feature \(d_{i-1}\) is first processed by a Fourier Feature Convolution (FFC) block and upsampled via a convolution. The result is then refined by a gated cross-attention module that integrates the bottleneck context \(b\)\cite{WOS:001213811603043}, and fused with the corresponding encoder feature \(e_{4-i}\) through skip connection and another FFC block, as follows:
\begin{equation}
d_i'' = \mathrm{FFC}\left( \left[ \mathrm{Attn}\left( \mathrm{UpConv}(\mathrm{FFC}(d_{i-1})),\, b \right),\ e_{4-i} \right] \right)
\label{eq:decoder_stage}
\end{equation}

This process allows the decoder to progressively restore spatial details while adaptively leveraging the global context from the bottleneck through cross-attention. The skip connections ensure that fine-grained encoder features are preserved and fused at each stage for richer reconstruction\cite{10.1007/978-3-319-24574-4_28}.

\subsection{Evidential Output and Losses}

The final output is produced by a \(3 \times 3\) convolution yielding four channels \((\gamma, v, \alpha, \beta)\), with Softplus activation to ensure positivity. The evidential regression loss combines a negative log-likelihood term \(\mathrm{NLL}(\gamma, v, \alpha, \beta; Y)\) and a regularization weighted by \(\lambda_{\mathrm{ev}}\) that penalizes deviations from the target \(Y\)\cite{NEURIPS2020_aab08546}:
\begin{equation}
\mathcal{L}_{\mathrm{ev}} = \mathrm{NLL}(\gamma, v, \alpha, \beta; Y) + \lambda_{\mathrm{ev}} |Y - \gamma| (2v + \alpha)
\label{eq:evidential_loss}
\end{equation}

Besides \(\mathcal{L}_{\mathrm{ev}}\), the overall generator loss \(\mathcal{L}_G\) integrates adversarial \(\mathcal{L}_{\mathrm{adv}}\), high-resolution perceptual \(\mathcal{L}_{\mathrm{HRF}}\), feature-matching \(\mathcal{L}_{\mathrm{FM}}\), and \(L_1\) reconstruction losses, weighted by \(\lambda\)-coefficients to encourage accurate, perceptually coherent outputs while maintaining uncertainty awareness:
\begin{equation}
\mathcal{L}_G = \sum_{i \in \{\mathrm{ev},\, \mathrm{adv},\, \mathrm{HRF},\, \mathrm{FM},\, 1\}} \lambda_i \mathcal{L}_i
\label{eq:total_loss}
\end{equation}

\section{Experiments}
\label{sec:experiments}

\subsection{Setup}

We employ a 5-layer PatchDiscriminator to compute the adversarial loss \(\mathcal{L}_{\mathrm{adv}}\) along with an R1 gradient penalty for stable training. The model is optimized using AdamW with a learning rate of \(10^{-4}\), applying cosine annealing for the generator and exponential decay for the discriminator. Training is performed with a batch size of 3. To ensure reproducibility and monitor progress, metrics including PSNR, SSIM, LPIPS, and FID are evaluated regularly~\cite{2018The,NIPS2017_8a1d6947}, and model checkpoints are saved every 5 epochs.

\textbf{Dataset and Splits.} Atmos-Bench contains 384 volumetric scenes simulated at two wavelengths (532\,nm and 355\,nm). For each wavelength, 90\% of the data is used for training and 10\% for testing.

\textbf{Baselines.} We benchmark our method against six state-of-the-art image restoration models—CAT~\cite{chen2023crossaggregationtransformerimage}, DDS2M~\cite{miao2023dds2mselfsuperviseddenoisingdiffusion}, EchoIR~\cite{he2024echoiradvancingimagerestoration}, TSFormer~\cite{10.1145/3696452}, UIR-LoRA~\cite{zhang2024uirloraachievinguniversalimage}, and VmambaIR~\cite{shi2024vmambairvisualstatespace}—on the Standard 3D Atmospheric Structure Dataset, covering both 532 nm and 355 nm subsets. All methods are evaluated under the same masking settings as FourCastX, and their BC reconstructions are assessed by PSNR, SSIM, MAE, LPIPS, and FID.

\textbf{Metrics.} We report five complementary evaluation metrics: PSNR, SSIM, MAE, LPIPS, and FID.

\textbf{Training.} All models are trained for 20 epochs using the AdamW optimizer with a learning rate of \(1 \times 10^{-4}\) and a batch size of 3. Cosine annealing is adopted for the generator learning rate schedule, while an exponential decay schedule is applied for the discriminator.

\textbf{Hardware.} All experiments are conducted on a single NVIDIA RTX 4090 GPU with 24 GB of memory. Training and inference on the full 3D volumes are performed without gradient checkpointing or model parallelism.

\subsection{Qualitative Comparison}

\begin{figure*}[ht]
  \centering
  \includegraphics[width=1\linewidth]{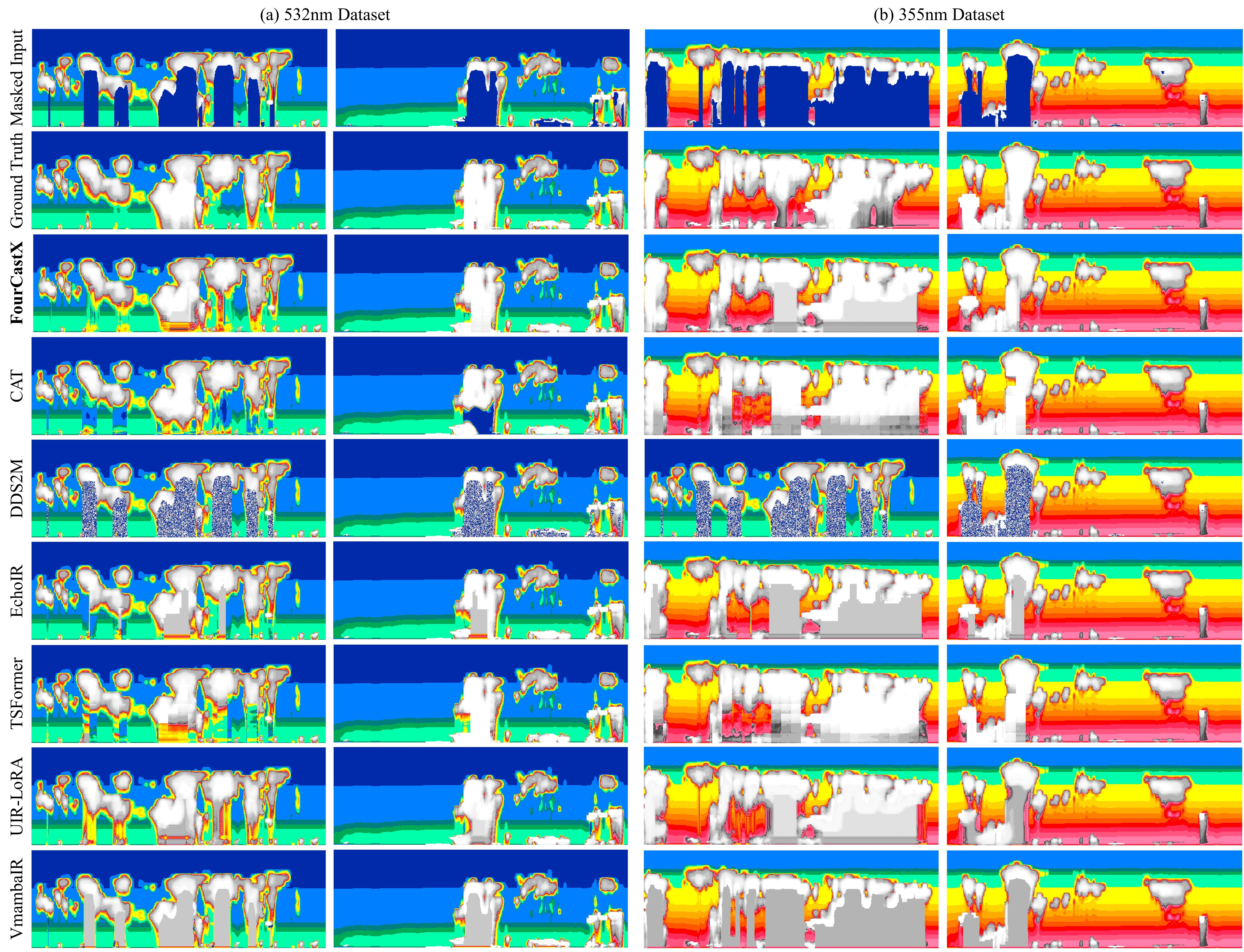}
  \caption{Visual results on (a) 532 nm and (b) 355 nm. From top to bottom: masked ATB input, ground truth BC, and model outputs of FourCastX, CAT, DDS2M, EchoIR, TSFormer, UIR-LoRA, and VmambaIR. Each dataset shows a heavy-mask example on the left and a light-mask example on the right, demonstrating model robustness under varying occlusion levels. All panels use the same colormap.}
  \label{fig:qualitative}
\end{figure*}

Figure~\ref{fig:qualitative} presents a side-by-side comparison on (a) 532 nm and (b) 355 nm across eight methods, evaluated by \textbf{pattern recovery}, \textbf{feature similarity}, and \textbf{restoration quality}, with additional emphasis on \textbf{noise suppression} and \textbf{texture continuity}.

\textbf{Pattern Recovery.} FourCastX excels at reconstructing thin, vertically elongated cloud filaments and sharp gradient transitions that TSFormer and UIR‑LoRA blur. DDS2M fails to recover these fine structures and instead produces scattered noise, while EchoIR oversimplifies and omits subtle features. CAT and VmambaIR capture coarse outlines but suffer from edge softening, resulting in less precise boundaries.

\textbf{Feature Similarity.} FourCastX outputs align tightly with ground truth in high-contrast regions (e.g., cloud tops), preserving boundary sharpness and structural integrity. By contrast, DDS2M’s speckle noise degrades fidelity, and CAT’s mild edge rounding reduces alignment. EchoIR yields overly smooth contours, and VmambaIR, though accurate at large scales, misses thin-layer details.

\textbf{Restoration Quality.} FourCastX achieves seamless inpainting, integrating restored areas without blotchy or granular artifacts. DDS2M shows pronounced speckle, EchoIR oversmooths textures, CAT occasionally over‑smooths edges, and TSFormer and UIR‑LoRA leave visible gaps in occluded regions.

\textbf{Noise Suppression.} FourCastX produces clean, artifact-free reconstructions. DDS2M exhibits pervasive speckle noise, EchoIR and CAT reduce noise at the cost of detail, and VmambaIR introduces low-level background artifacts, especially in the 355 nm subset.

\textbf{Texture Continuity.} FourCastX preserves high-frequency cloud internal structures while maintaining volumetric coherence. DDS2M’s outputs appear patchy, EchoIR fails to reproduce fine textures, and TSFormer and UIR‑LoRA fragment textures near cloud edges.

\textbf{Cross-Dataset Robustness.} FourCastX maintains its advantages on both 532 nm (realistic backscatter) and 355 nm (idealized aerosol) data. Other methods (e.g., DDS2M, TSFormer) degrade on the more challenging 355 nm subset, whereas FourCastX retains structural accuracy and boundary clarity, demonstrating superior generalization under varying input modalities.

Overall, FourCastX outperforms all baselines in recovering cloud structures with high fidelity, minimal artifacts, and robust texture preservation. These qualitative gains are supported by quantitative improvements in SSIM, PSNR, and other metrics reported in subsequent sections.

\subsection{Quantitative Results}

Table~\ref{tab:method-comparison} shows that FourCastX achieves a PSNR of 23.38 dB at 532 nm—over 6 dB higher than the strongest baseline—and 23.94 dB at 355 nm, marking improvements of 23\% and 15\%, respectively. Similarly, its SSIM scores of 0.969 and 0.970 represent gains of 7\% and 5\% over the next best models. In both bands, FourCastX reduces MAE to under 0.01—an 92\% to 71\% decrease—and achieves the lowest LPIPS and FID, confirming superior pixel-level accuracy and perceptual realism.

These consistent gains across metrics and datasets demonstrate FourCastX’s robustness in reconstructing fine-scale volumetric details and maintaining global consistency under heavy masking. The improvements in perceptual metrics (LPIPS and FID) further indicate enhanced realism, which is critical for downstream climate analysis and simulation fidelity.

Moreover, FourCastX consistently outperforms across wavelengths: at 532 nm it cuts MAE by over 70\% and lowers FID to 28.02, while at 355 nm it reduces MAE by 70\% with an FID of 25.46. These results underscore its robustness under heavy masking and its ability to recover both fine‑scale cloud filaments and broad volumetric structures without auxiliary inputs.

\definecolor{myred}{RGB}{215,94,128}
\definecolor{mygreen}{RGB}{86,188,80}

\newcommand{\incsmall}[1]{\hspace{-0.02em}\scalebox{0.8}{\textcolor{myred}{\ding{115}#1\%}}}
\newcommand{\decsmall}[1]{\hspace{-0.02em}\scalebox{0.8}{\textcolor{mygreen}{\ding{116}#1\%}}}

\begin{table*}[ht]
\centering
\caption{Performance comparison of state-of-the-art methods on two wavelengths.}
\label{tab:method-comparison}  
\begin{threeparttable}
\renewcommand{\arraystretch}{1}
\setlength{\tabcolsep}{0pt}  
\begin{tabular}{
  >{\centering\arraybackslash}m{1.8cm}  
  *{5}{>{\raggedright\arraybackslash}m{1.6cm}}  
  *{5}{>{\raggedright\arraybackslash}m{1.6cm}}  
}
\toprule

Dataset & \multicolumn{5}{c}{532nm} & \multicolumn{5}{c}{355nm} \\
\cmidrule(lr){2-6} \cmidrule(lr){7-11}
Method  & PSNR \(\uparrow\) & SSIM \(\uparrow\) & MAE \(\downarrow\) & LPIPS \(\downarrow\) & FID \(\downarrow\) &
         PSNR \(\uparrow\) & SSIM \(\uparrow\) & MAE \(\downarrow\) & LPIPS \(\downarrow\) & FID \(\downarrow\) \\              
\midrule
\textbf{FourCastX} &
\textbf{23.38} & \textbf{0.969} & \textbf{0.008} & \textbf{0.032} & \textbf{28.02} &
\textbf{23.94} & \textbf{0.970} & \textbf{0.006} & \textbf{0.027} & \textbf{25.46} \\

CAT & 
17.04\incsmall{37} & 0.807\incsmall{20} & 0.046\incsmall{83} & 0.224\incsmall{86} & 181.6\incsmall{85} &
12.72\incsmall{88} & 0.491\incsmall{98} & 0.089\incsmall{93} & 0.343\incsmall{92} & 480.4\incsmall{95} \\

DDS2M & 
12.47\incsmall{87} & 0.808\incsmall{20} & 0.074\incsmall{89} & 0.372\incsmall{91} & 152.5\incsmall{82} &
13.94\incsmall{72} & 0.847\incsmall{15} & 0.056\incsmall{89} & 0.299\incsmall{91} & 153.0\incsmall{83} \\

EchoIR &
18.27\incsmall{28} & 0.902\incsmall{7} & 0.034\incsmall{76} & 0.135\incsmall{76} & 62.51\incsmall{55} &
19.13\incsmall{25} & 0.924\incsmall{5} & 0.029\incsmall{79} & 0.107\incsmall{75} & 55.43\incsmall{54} \\

TSFormer &
18.78\incsmall{24} & 0.877\incsmall{10} & 0.028\incsmall{71} & 0.179\incsmall{82} & 59.32\incsmall{53} &
20.81\incsmall{15} & 0.927\incsmall{5} & 0.021\incsmall{71} & 0.151\incsmall{82} & 49.41\incsmall{48} \\

UIR-LoRA &
18.94\incsmall{23} & 0.890\incsmall{9} & 0.032\incsmall{75} & 0.129\incsmall{75} & 40.86\incsmall{31} &
20.01\incsmall{20} & 0.924\incsmall{5} & 0.025\incsmall{76} & 0.096\incsmall{72} & 37.51\incsmall{32} \\

VmambaIR &
16.75\incsmall{40} & 0.864\incsmall{12} & 0.046\incsmall{83} & 0.172\incsmall{81} & 66.09\incsmall{58} &
17.92\incsmall{34} & 0.899\incsmall{8} & 0.034\incsmall{82} & 0.150\incsmall{82} & 60.98\incsmall{58} \\

\bottomrule
\end{tabular}
\end{threeparttable}
\end{table*}

\subsection{Strengths}

\textbf{Superior reconstruction quality.} As Table~\ref{tab:method-comparison} demonstrates, our method leads all baselines by over 2 dB in PSNR and 0.02 in SSIM at both 532 nm and 355 nm, while also achieving the lowest MAE, LPIPS, and FID scores \cite{Zhang_2018_CVPR}. This confirms its ability to recover thin layers and sharp boundaries with high fidelity.

\textbf{Efficient capacity.} Despite its performance, our generator has only one-quarter the parameters of CoModGAN and one-third those of MADF \cite{9440708}, thanks to dynamic expert routing via Gated MoE \cite{shazeer2017outrageouslylargeneuralnetworks}. This compact design enables fast convergence and reduced memory footprint.

\textbf{Physics‐aware stability.} By embedding the ATB–BC energy relation as a differentiable constraint, our model preserves physical consistency under heavy occlusion, handling multiple scattering without auxiliary inputs. As mask rates increase, performance degrades by less than 5\%, outperforming purely data‐driven methods.

\section{Conclusion and Future Work}
\label{Conclusion}

In this work, we presented FourCastX, a unified physics-aware generative framework for 3D atmospheric backscatter retrieval. Our method integrates high-fidelity WRF–COSP simulations with standardized evaluation protocols, establishing a new benchmark for this challenging task. FourCastX consistently outperforms classical inversion methods and recent learning-based approaches on synthetic datasets, demonstrating strong reconstruction accuracy with significantly fewer parameters than competing models~\cite{doi:10.1137/18M1225409}. Future work will focus on extending our framework to real-world satellite observations and further improving the model’s generalization capabilities.

\paragraph{Limitations}  
Atmos-Bench’s reliance on WRF–COSP inherits biases from microphysical and radiative schemes \cite{https://doi.org/10.1029/2019JD031413,gmd-16-199-2023,WOS:001414960500001}, potentially limiting generalization to unmodeled conditions. The current design targets single-channel LiDAR backscatter; multi-wavelength or multi-sensor fusion (e.g., SAR, passive radiometry) remains to be explored \cite{rs9090890}. The Gated MoE can exhibit early-stage routing instability, necessitating robust balancing strategies \cite{wang2024auxiliarylossfreeloadbalancingstrategy,royer2023revisitingsinglegatedmixturesexperts,chen2022understandingmixtureexpertsdeep}. Finally, as models approach the aleatoric uncertainty floor of ATB measurements, finer-grained evaluation and uncertainty quantification will be required \cite{amt-15-149-2022}.

\paragraph{Future Directions}  
To enhance Atmos-Bench, future work could: (1) incorporate semi-supervised fine-tuning on diverse real LiDAR datasets to mitigate simulator–reality gaps; (2) extend to multi-sensor fusion for richer volumetric reconstructions; (3) develop lightweight FFC and attention variants to accelerate full-volume inference; and (4) investigate uncertainty-driven training—e.g., auxiliary load‑balancing losses or Bayesian MoE—to stabilize expert routing and approach theoretical performance limits.

By providing a unified, high-fidelity dataset, standardized evaluation protocols, and a strong physics-informed generative baseline, Atmos-Bench establishes a foundational platform for next-generation machine learning models in atmospheric science. This effort aims to empower researchers and practitioners to deliver more precise, reliable, and actionable insights, ultimately contributing to improved climate monitoring, weather forecasting, and environmental decision-making in the face of global change.


\bibliography{aaai25}

\begin{thebibliography}{73}
\providecommand{\natexlab}[1]{#1}

\bibitem[{Adomako et~al.(2024)Adomako, Jamshidi, Yusup, Elsebakhi, Jaafar,
  Ishak, Lim, and Ahmad}]{WOS:001414960500001}
Adomako, A.~B.; Jamshidi, E.~J.; Yusup, Y.; Elsebakhi, E.; Jaafar, M.~H.;
  Ishak, M. I.~S.; Lim, H.~S.; and Ahmad, M.~I. 2024.
\newblock Deep learning approaches for bias correction in WRF model outputs for
  enhanced solar and wind energy estimation: A case study in East and West
  Malaysia.
\newblock \emph{ECOLOGICAL INFORMATICS}, 84.

\bibitem[{Ahn et~al.(2023)Ahn, Kim, Hong, and Ko}]{WOS:000971500203043}
Ahn, D.; Kim, S.; Hong, H.; and Ko, B.~C. 2023.
\newblock STAR-Transformer: A Spatio-temporal Cross Attention Transformer for
  Human Action Recognition.
\newblock In \emph{2023 IEEE/CVF WINTER CONFERENCE ON APPLICATIONS OF COMPUTER
  VISION (WACV)}, IEEE Winter Conference on Applications of Computer Vision,
  3319--3328. IEEE; CVF; IEEE Comp Soc.
\newblock ISBN 978-1-6654-9346-8.
\newblock 23rd IEEE/CVF Winter Conference on Applications of Computer Vision
  (WACV), Waikoloa, HI, JAN 03-07, 2023.

\bibitem[{Alkin et~al.(2024)Alkin, Beck, P{\"o}ppel, Hochreiter, and
  Brandstetter}]{alkin2024vision}
Alkin, B.; Beck, M.; P{\"o}ppel, K.; Hochreiter, S.; and Brandstetter, J. 2024.
\newblock Vision-lstm: xlstm as generic vision backbone.
\newblock \emph{arXiv preprint arXiv:2406.04303}.

\bibitem[{Amini et~al.(2020)Amini, Schwarting, Soleimany, and
  Rus}]{NEURIPS2020_aab08546}
Amini, A.; Schwarting, W.; Soleimany, A.; and Rus, D. 2020.
\newblock Deep Evidential Regression.
\newblock In Larochelle, H.; Ranzato, M.; Hadsell, R.; Balcan, M.; and Lin, H.,
  eds., \emph{Advances in Neural Information Processing Systems}, volume~33,
  14927--14937. Curran Associates, Inc.

\bibitem[{Bonazzola et~al.(2023)Bonazzola, Chepfer, Ma, Quaas, Winker,
  Feofilov, and Schutgens}]{gmd-16-1359-2023}
Bonazzola, M.; Chepfer, H.; Ma, P.-L.; Quaas, J.; Winker, D.~M.; Feofilov, A.;
  and Schutgens, N. 2023.
\newblock Incorporation of aerosol into the COSPv2 satellite lidar simulator
  for climate model evaluation.
\newblock \emph{Geoscientific Model Development}, 16(4): 1359--1377.

\bibitem[{Bountos et~al.(2025)Bountos, Ouaknine, Papoutsis, and
  Rolnick}]{bountos2025fomomultimodalmultiscalemultitask}
Bountos, N.~I.; Ouaknine, A.; Papoutsis, I.; and Rolnick, D. 2025.
\newblock FoMo: Multi-Modal, Multi-Scale and Multi-Task Remote Sensing
  Foundation Models for Forest Monitoring.
\newblock arXiv:2312.10114.

\bibitem[{Chen et~al.(2022{\natexlab{a}})Chen, Deng, Wu, Gu, and
  Li}]{chen2022understandingmixtureexpertsdeep}
Chen, Z.; Deng, Y.; Wu, Y.; Gu, Q.; and Li, Y. 2022{\natexlab{a}}.
\newblock Towards Understanding Mixture of Experts in Deep Learning.
\newblock arXiv:2208.02813.

\bibitem[{Chen et~al.(2022{\natexlab{b}})Chen, Deng, Wu, Gu, and
  Li}]{WOS:001213927509018}
Chen, Z.; Deng, Y.; Wu, Y.; Gu, Q.; and Li, Y. 2022{\natexlab{b}}.
\newblock Towards Understanding the Mixture-of-Experts Layer in Deep Learning.
\newblock In Koyejo, S.; Mohamed, S.; Agarwal, A.; Belgrave, D.; Cho, K.; and
  Oh, A., eds., \emph{ADVANCES IN NEURAL INFORMATION PROCESSING SYSTEMS 35
  (NEURIPS 2022)}, Advances in Neural Information Processing Systems.
\newblock ISBN 978-1-7138-7108-8.
\newblock 36th Conference on Neural Information Processing Systems (NeurIPS),
  ELECTR NETWORK, NOV 28-DEC 09, 2022.

\bibitem[{Chen et~al.(2022{\natexlab{c}})Chen, Zhang, Gu, Zhang, Kong, and
  Yuan}]{zheng2022cross}
Chen, Z.; Zhang, Y.; Gu, J.; Zhang, Y.; Kong, L.; and Yuan, X.
  2022{\natexlab{c}}.
\newblock Cross Aggregation Transformer for Image Restoration.
\newblock In Oh, A.~H.; Agarwal, A.; Belgrave, D.; and Cho, K., eds.,
  \emph{Advances in Neural Information Processing Systems}.

\bibitem[{Chen et~al.(2023)Chen, Zhang, Gu, Zhang, Kong, and
  Yuan}]{chen2023crossaggregationtransformerimage}
Chen, Z.; Zhang, Y.; Gu, J.; Zhang, Y.; Kong, L.; and Yuan, X. 2023.
\newblock Cross Aggregation Transformer for Image Restoration.
\newblock arXiv:2211.13654.

\bibitem[{Chi, Jiang, and Mu(2020)}]{Chi_2020_FFC}
Chi, L.; Jiang, B.; and Mu, Y. 2020.
\newblock Fast Fourier Convolution.
\newblock In \emph{Advances in Neural Information Processing Systems}.

\bibitem[{Cho et~al.(2020)Cho, Jun, Ho, and
  McFarquhar}]{https://doi.org/10.1029/2019JD031413}
Cho, H.; Jun, S.-Y.; Ho, C.-H.; and McFarquhar, G. 2020.
\newblock Simulations of Winter Arctic Clouds and Associated Radiation Fluxes
  Using Different Cloud Microphysics Schemes in the Polar WRF: Comparisons With
  CloudSat, CALIPSO, and CERES.
\newblock \emph{Journal of Geophysical Research: Atmospheres}, 125(2):
  e2019JD031413.
\newblock E2019JD031413 2019JD031413.

\bibitem[{Clarke and
  Kim(2005)}]{TheResponseTimeoftheTemperatureoftheEquatorialTropospheretoENSOHeating}
Clarke, A.~J.; and Kim, K.-Y. 2005.
\newblock The Response Time of the Temperature of the Equatorial Troposphere to
  ENSO Heating.
\newblock \emph{Journal of the Atmospheric Sciences}, 62(12): 4412 -- 4422.

\bibitem[{Cong et~al.(2023)Cong, Khanna, Meng, Liu, Rozi, He, Burke, Lobell,
  and Ermon}]{cong2023satmaepretrainingtransformerstemporal}
Cong, Y.; Khanna, S.; Meng, C.; Liu, P.; Rozi, E.; He, Y.; Burke, M.; Lobell,
  D.~B.; and Ermon, S. 2023.
\newblock SatMAE: Pre-training Transformers for Temporal and Multi-Spectral
  Satellite Imagery.
\newblock arXiv:2207.08051.

\bibitem[{Dong et~al.(2016)Dong, Loy, He, and Tang}]{7115171}
Dong, C.; Loy, C.~C.; He, K.; and Tang, X. 2016.
\newblock Image Super-Resolution Using Deep Convolutional Networks.
\newblock \emph{IEEE Transactions on Pattern Analysis and Machine
  Intelligence}, 38(2): 295--307.

\bibitem[{Donovan, van Zadelhoff, and Wang(2024)}]{amt-17-5301-2024}
Donovan, D.~P.; van Zadelhoff, G.-J.; and Wang, P. 2024.
\newblock The EarthCARE lidar cloud and aerosol profile processor (A-PRO): the
  A-AER, A-EBD, A-TC, and A-ICE products.
\newblock \emph{Atmospheric Measurement Techniques}, 17(17): 5301--5340.

\bibitem[{Fuller, Millard, and Green(2023)}]{NEURIPS2023_11822e84}
Fuller, A.; Millard, K.; and Green, J. 2023.
\newblock CROMA: Remote Sensing Representations with Contrastive Radar-Optical
  Masked Autoencoders.
\newblock In Oh, A.; Naumann, T.; Globerson, A.; Saenko, K.; Hardt, M.; and
  Levine, S., eds., \emph{Advances in Neural Information Processing Systems},
  volume~36, 5506--5538. Curran Associates, Inc.

\bibitem[{Guo et~al.(2025)Guo, Guo, Zha, Zhang, Li, Dai, Xia, and
  Li}]{guo2025mambairv2attentivestatespace}
Guo, H.; Guo, Y.; Zha, Y.; Zhang, Y.; Li, W.; Dai, T.; Xia, S.-T.; and Li, Y.
  2025.
\newblock MambaIRv2: Attentive State Space Restoration.
\newblock arXiv:2411.15269.

\bibitem[{Guo et~al.(2023)Guo, Lao, Dang, Zhang, Yu, Ru, Zhong, Huang, Wu, and
  Hu}]{2023SkySense}
Guo, X.; Lao, J.; Dang, B.; Zhang, Y.; Yu, L.; Ru, L.; Zhong, L.; Huang, Z.;
  Wu, K.; and Hu, D. 2023.
\newblock SkySense: A Multi-Modal Remote Sensing Foundation Model Towards
  Universal Interpretation for Earth Observation Imagery.
\newblock \emph{IEEE}.

\bibitem[{Guo et~al.(2024)Guo, Lao, Dang, Zhang, Yu, Ru, Zhong, Huang, Wu, Hu,
  He, Wang, Chen, Yang, Zhang, and Li}]{guo2024skysensemultimodalremotesensing}
Guo, X.; Lao, J.; Dang, B.; Zhang, Y.; Yu, L.; Ru, L.; Zhong, L.; Huang, Z.;
  Wu, K.; Hu, D.; He, H.; Wang, J.; Chen, J.; Yang, M.; Zhang, Y.; and Li, Y.
  2024.
\newblock SkySense: A Multi-Modal Remote Sensing Foundation Model Towards
  Universal Interpretation for Earth Observation Imagery.
\newblock arXiv:2312.10115.

\bibitem[{Han et~al.(2025)Han, Nguyen, Harris, Ho, and
  Saria}]{han2025fusemoemixtureofexpertstransformersfleximodal}
Han, X.; Nguyen, H.; Harris, C.; Ho, N.; and Saria, S. 2025.
\newblock FuseMoE: Mixture-of-Experts Transformers for Fleximodal Fusion.
\newblock arXiv:2402.03226.

\bibitem[{He(2024)}]{he2024epl}
He, Y. 2024.
\newblock Epl: Evidential prototype learning for semi-supervised medical image
  segmentation.
\newblock \emph{arXiv preprint arXiv:2404.06181}.

\bibitem[{He et~al.(2024{\natexlab{a}})He, Bi, Li, Pun, Jiao, and
  Jin}]{he2024mutual}
He, Y.; Bi, Y.; Li, L.; Pun, C.-M.; Jiao, W.; and Jin, Z. 2024{\natexlab{a}}.
\newblock Mutual evidential deep learning for semi-supervised medical image
  segmentation.
\newblock In \emph{2024 IEEE International Conference on Bioinformatics and
  Biomedicine (BIBM)}, 2010--2017. IEEE.

\bibitem[{He and He(2024)}]{he2024echoiradvancingimagerestoration}
He, Y.; and He, Y. 2024.
\newblock EchoIR: Advancing Image Restoration with Echo Upsampling and Bi-Level
  Optimization.
\newblock arXiv:2412.07225.

\bibitem[{He and Li(2024)}]{he2024uncertainty}
He, Y.; and Li, L. 2024.
\newblock Uncertainty-aware evidential fusion-based learning for
  semi-supervised medical image segmentation.
\newblock \emph{arXiv preprint arXiv:2404.06177}.

\bibitem[{He et~al.(2024{\natexlab{b}})He, Li, Zhan, Jiao, and
  Pun}]{he2024generalized}
He, Y.; Li, L.; Zhan, T.; Jiao, W.; and Pun, C.-M. 2024{\natexlab{b}}.
\newblock Generalized uncertainty-based evidential fusion with hybrid
  multi-head attention for weak-supervised temporal action localization.
\newblock In \emph{ICASSP 2024-2024 IEEE International Conference on Acoustics,
  Speech and Signal Processing (ICASSP)}, 3855--3859. IEEE.

\bibitem[{He et~al.(2025)He, Li, Zhan, Pun, Jiao, and Jin}]{he2025co}
He, Y.; Li, L.; Zhan, T.; Pun, C.-M.; Jiao, W.; and Jin, Z. 2025.
\newblock Co-evidential fusion with information volume for semi-supervised
  medical image segmentation.
\newblock \emph{Pattern Recognition}, 166: 111639.

\bibitem[{He et~al.(2024{\natexlab{c}})He, Song, Li, Zhan, and
  Jiao}]{he2024residual}
He, Y.; Song, W.; Li, L.; Zhan, T.; and Jiao, W. 2024{\natexlab{c}}.
\newblock Residual Feature-Reutilization Inception Network.
\newblock \emph{Pattern Recognition}, 110439.

\bibitem[{Heusel et~al.(2017)Heusel, Ramsauer, Unterthiner, Nessler, and
  Hochreiter}]{NIPS2017_8a1d6947}
Heusel, M.; Ramsauer, H.; Unterthiner, T.; Nessler, B.; and Hochreiter, S.
  2017.
\newblock GANs Trained by a Two Time-Scale Update Rule Converge to a Local Nash
  Equilibrium.
\newblock In Guyon, I.; Luxburg, U.~V.; Bengio, S.; Wallach, H.; Fergus, R.;
  Vishwanathan, S.; and Garnett, R., eds., \emph{Advances in Neural Information
  Processing Systems}, volume~30. Curran Associates, Inc.

\bibitem[{Hong et~al.(2024{\natexlab{a}})Hong, Zhang, Li, Li, Li, Yao, Yokoya,
  Li, Ghamisi, Jia, Plaza, Gamba, Benediktsson, and Chanussot}]{10490262}
Hong, D.; Zhang, B.; Li, X.; Li, Y.; Li, C.; Yao, J.; Yokoya, N.; Li, H.;
  Ghamisi, P.; Jia, X.; Plaza, A.; Gamba, P.; Benediktsson, J.~A.; and
  Chanussot, J. 2024{\natexlab{a}}.
\newblock { SpectralGPT: Spectral Remote Sensing Foundation Model }.
\newblock \emph{IEEE Transactions on Pattern Analysis \& Machine Intelligence},
  46(08): 5227--5244.

\bibitem[{Hong et~al.(2024{\natexlab{b}})Hong, Zhang, Li, Li, Li, Yao, Yokoya,
  Li, Ghamisi, Jia, Plaza, Gamba, Benediktsson, and
  Chanussot}]{10.1109/TPAMI.2024.3362475}
Hong, D.; Zhang, B.; Li, X.; Li, Y.; Li, C.; Yao, J.; Yokoya, N.; Li, H.;
  Ghamisi, P.; Jia, X.; Plaza, A.; Gamba, P.; Benediktsson, J.~A.; and
  Chanussot, J. 2024{\natexlab{b}}.
\newblock SpectralGPT: Spectral Remote Sensing Foundation Model.
\newblock \emph{IEEE Trans. Pattern Anal. Mach. Intell.}, 46(8): 5227–5244.

\bibitem[{Hu and Lu(2024)}]{10.3389/frsen.2024.1477503}
Hu, K.; and Lu, X. 2024.
\newblock Spaceborne lidar measurement of global cloud properties through
  machine learning.
\newblock \emph{Frontiers in Remote Sensing}, Volume 5 - 2024.

\bibitem[{Irwin et~al.(2017)Irwin, Beaulne, Braun, and Fotopoulos}]{rs9090890}
Irwin, K.; Beaulne, D.; Braun, A.; and Fotopoulos, G. 2017.
\newblock Fusion of SAR, Optical Imagery and Airborne LiDAR for Surface Water
  Detection.
\newblock \emph{Remote Sensing}, 9(9).

\bibitem[{Jain et~al.(2024)Jain, Hegde, Kusupati, Nagrani, Buch, Jain, Arnab,
  and Paul}]{jain2024mixturenestedexpertsadaptive}
Jain, G.; Hegde, N.; Kusupati, A.; Nagrani, A.; Buch, S.; Jain, P.; Arnab, A.;
  and Paul, S. 2024.
\newblock Mixture of Nested Experts: Adaptive Processing of Visual Tokens.
\newblock arXiv:2407.19985.

\bibitem[{Johnson, Alahi, and Fei-Fei(2016)}]{WOS:000389383900043}
Johnson, J.; Alahi, A.; and Fei-Fei, L. 2016.
\newblock Perceptual Losses for Real-Time Style Transfer and Super-Resolution.
\newblock In Leibe, B.; Matas, J.; Sebe, N.; and Welling, M., eds.,
  \emph{COMPUTER VISION - ECCV 2016, PT II}, volume 9906 of \emph{Lecture Notes
  in Computer Science}, 694--711.
\newblock ISBN 978-3-319-46475-6; 978-3-319-46474-9.
\newblock 14th European Conference on Computer Vision (ECCV), Amsterdam,
  NETHERLANDS, OCT 08-16, 2016.

\bibitem[{Lacoste et~al.(2023)Lacoste, Lehmann, Rodriguez, Sherwin, Kerner,
  L\"{u}tjens, Irvin, Dao, Alemohammad, Drouin, Gunturkun, Huang, Vazquez,
  Newman, Bengio, Ermon, and Zhu}]{NEURIPS2023_a0644215}
Lacoste, A.; Lehmann, N.; Rodriguez, P.; Sherwin, E.; Kerner, H.; L\"{u}tjens,
  B.; Irvin, J.; Dao, D.; Alemohammad, H.; Drouin, A.; Gunturkun, M.; Huang,
  G.; Vazquez, D.; Newman, D.; Bengio, Y.; Ermon, S.; and Zhu, X. 2023.
\newblock GEO-Bench: Toward Foundation Models for Earth Monitoring.
\newblock In Oh, A.; Naumann, T.; Globerson, A.; Saenko, K.; Hardt, M.; and
  Levine, S., eds., \emph{Advances in Neural Information Processing Systems},
  volume~36, 51080--51093. Curran Associates, Inc.

\bibitem[{Li et~al.(2023)Li, Su, Yang, Jiang, Wang, and
  Xu}]{WOS:001275019903043}
Li, J.; Su, Q.; Yang, Y.; Jiang, Y.; Wang, C.; and Xu, H. 2023.
\newblock Adaptive Gating in Mixture-of-Experts based Language Models.
\newblock In Bouamor, H.; Pino, J.; and Bali, K., eds., \emph{2023 CONFERENCE
  ON EMPIRICAL METHODS IN NATURAL LANGUAGE PROCESSING, EMNLP 2023}, 3577--3587.
  Apple; Colossal AI; Google Res; GTCOM; King Salman Global Acad Arabic
  Language; LivePerson; SONY; Ahrefs; Alibaba Cloud; Amazon Sci; Baidu;
  ByteDance; Cohere; Megagon Labs; NEC; ANT Grp; Bloomberg Engn; HUAWEI; J P
  Morgan Chase \& Co; Salesforce; SAP; AiXplain; Duolingo; Jenni; Translated;
  Adobe; Babelscape; ModelBest; Nyonic; Mercari.
\newblock ISBN 979-8-89176-060-8.
\newblock Conference on Empirical Methods in Natural Language Processing
  (EMNLP), Singapore, SINGAPORE, DEC 06-10, 2023.

\bibitem[{Lin and Wang(2024)}]{10.1145/3696452}
Lin, J.; and Wang, Y.-G. 2024.
\newblock TSFormer: Tracking Structure Transformer for Image Inpainting.
\newblock \emph{ACM Trans. Multimedia Comput. Commun. Appl.}, 20(12).

\bibitem[{Liu et~al.(2021)Liu, Guo, Zhang, Zhang, Guan, and Xu}]{atmos12040481}
Liu, C.; Guo, J.; Zhang, B.; Zhang, H.; Guan, P.; and Xu, R. 2021.
\newblock A Reliability Assessment of the NCEP/FNL Reanalysis Data in Depicting
  Key Meteorological Factors on Clean Days and Polluted Days in Beijing.
\newblock \emph{Atmosphere}, 12(4).

\bibitem[{Liu et~al.(2024)Liu, Chen, Guan, Zhou, Zhu, Ye, Fu, and
  Zhou}]{liu2024remoteclipvisionlanguagefoundation}
Liu, F.; Chen, D.; Guan, Z.; Zhou, X.; Zhu, J.; Ye, Q.; Fu, L.; and Zhou, J.
  2024.
\newblock RemoteCLIP: A Vision Language Foundation Model for Remote Sensing.
\newblock arXiv:2306.11029.

\bibitem[{Liu et~al.(2022)Liu, Kim, Muzio, and
  Awadalla}]{liu2022gatingdropoutcommunicationefficientregularization}
Liu, R.; Kim, Y.~J.; Muzio, A.; and Awadalla, H.~H. 2022.
\newblock Gating Dropout: Communication-efficient Regularization for Sparsely
  Activated Transformers.
\newblock arXiv:2205.14336.

\bibitem[{Miao et~al.(2023{\natexlab{a}})Miao, Zhang, Zhang, and
  Tao}]{miao2023dds2m}
Miao, Y.; Zhang, L.; Zhang, L.; and Tao, D. 2023{\natexlab{a}}.
\newblock DDS2M: Self-Supervised Denoising Diffusion Spatio-Spectral Model for
  Hyperspectral Image Restoration.
\newblock \emph{arXiv preprint arXiv:2303.06682}.

\bibitem[{Miao et~al.(2023{\natexlab{b}})Miao, Zhang, Zhang, and
  Tao}]{miao2023dds2mselfsuperviseddenoisingdiffusion}
Miao, Y.; Zhang, L.; Zhang, L.; and Tao, D. 2023{\natexlab{b}}.
\newblock DDS2M: Self-Supervised Denoising Diffusion Spatio-Spectral Model for
  Hyperspectral Image Restoration.
\newblock arXiv:2303.06682.

\bibitem[{Nguyen et~al.(2022)Nguyen, Pham, Nguyen, Nguyen, Osher, and
  Ho}]{WOS:001213811603043}
Nguyen, T.~M.; Pham, M.; Nguyen, T.; Nguyen, K.; Osher, S.~J.; and Ho, N. 2022.
\newblock FourierFormer: Transformer Meets Generalized Fourier Integral
  Theorem.
\newblock In Koyejo, S.; Mohamed, S.; Agarwal, A.; Belgrave, D.; Cho, K.; and
  Oh, A., eds., \emph{ADVANCES IN NEURAL INFORMATION PROCESSING SYSTEMS 35
  (NEURIPS 2022)}, Advances in Neural Information Processing Systems.
\newblock ISBN 978-1-7138-7108-8.
\newblock 36th Conference on Neural Information Processing Systems (NeurIPS),
  ELECTR NETWORK, NOV 28-DEC 09, 2022.

\bibitem[{Ramirez-Jaime et~al.(2025)Ramirez-Jaime, Arce, Porras-Diaz,
  Ieremeiev, Rubel, Lukin, Kopytek, Lech, Fastowicz, and Okarma}]{rs17071215}
Ramirez-Jaime, A.; Arce, G.~R.; Porras-Diaz, N.; Ieremeiev, O.; Rubel, A.;
  Lukin, V.; Kopytek, M.; Lech, P.; Fastowicz, J.; and Okarma, K. 2025.
\newblock Generative Diffusion Models for Compressed Sensing of Satellite LiDAR
  Data: Evaluating Image Quality Metrics in Forest Landscape Reconstruction.
\newblock \emph{Remote Sensing}, 17(7).

\bibitem[{Ronneberger, Fischer, and Brox(2015)}]{10.1007/978-3-319-24574-4_28}
Ronneberger, O.; Fischer, P.; and Brox, T. 2015.
\newblock U-Net: Convolutional Networks for Biomedical Image Segmentation.
\newblock In Navab, N.; Hornegger, J.; Wells, W.~M.; and Frangi, A.~F., eds.,
  \emph{Medical Image Computing and Computer-Assisted Intervention -- MICCAI
  2015}, 234--241. Cham: Springer International Publishing.
\newblock ISBN 978-3-319-24574-4.

\bibitem[{Royer et~al.(2023)Royer, Karmanov, Skliar, Bejnordi, and
  Blankevoort}]{royer2023revisitingsinglegatedmixturesexperts}
Royer, A.; Karmanov, I.; Skliar, A.; Bejnordi, B.~E.; and Blankevoort, T. 2023.
\newblock Revisiting Single-gated Mixtures of Experts.
\newblock arXiv:2304.05497.

\bibitem[{Selmer et~al.(2024)Selmer, Yorks, Nowottnick, Cresanti, and
  Christian}]{rs16152735}
Selmer, P.; Yorks, J.~E.; Nowottnick, E.~P.; Cresanti, A.; and Christian, K.~E.
  2024.
\newblock A Deep Learning Lidar Denoising Approach for Improving Atmospheric
  Feature Detection.
\newblock \emph{Remote Sensing}, 16(15).

\bibitem[{Shazeer et~al.(2017)Shazeer, Mirhoseini, Maziarz, Davis, Le, Hinton,
  and Dean}]{shazeer2017outrageouslylargeneuralnetworks}
Shazeer, N.; Mirhoseini, A.; Maziarz, K.; Davis, A.; Le, Q.; Hinton, G.; and
  Dean, J. 2017.
\newblock Outrageously Large Neural Networks: The Sparsely-Gated
  Mixture-of-Experts Layer.
\newblock arXiv:1701.06538.

\bibitem[{Shi et~al.(2015)Shi, Chen, Wang, Yeung, Wong, and
  Woo}]{WOS:000450913102103}
Shi, X.; Chen, Z.; Wang, H.; Yeung, D.-Y.; Wong, W.-k.; and Woo, W.-c. 2015.
\newblock Convolutional LSTM Network: A Machine Learning Approach for
  Precipitation Nowcasting.
\newblock In Cortes, C.; Lawrence, N.; Lee, D.; Sugiyama, M.; and Garnett, R.,
  eds., \emph{ADVANCES IN NEURAL INFORMATION PROCESSING SYSTEMS 28 (NIPS
  2015)}, volume~28 of \emph{Advances in Neural Information Processing
  Systems}.
\newblock 29th Annual Conference on Neural Information Processing Systems
  (NIPS), Montreal, CANADA, DEC 07-12, 2015.

\bibitem[{Shi et~al.(2024)Shi, Xia, Jin, Wang, Zhao, Xia, Xiao, and
  Yang}]{shi2024vmambairvisualstatespace}
Shi, Y.; Xia, B.; Jin, X.; Wang, X.; Zhao, T.; Xia, X.; Xiao, X.; and Yang, W.
  2024.
\newblock VmambaIR: Visual State Space Model for Image Restoration.
\newblock arXiv:2403.11423.

\bibitem[{Sinha, Moorthi, and Dhar(2022)}]{WOS:000861612700053}
Sinha, A.~K.; Moorthi, S.~M.; and Dhar, D. 2022.
\newblock NL-FFC: Non-Local Fast Fourier Convolution for Image Super
  Resolution.
\newblock In \emph{2022 IEEE/CVF CONFERENCE ON COMPUTER VISION AND PATTERN
  RECOGNITION WORKSHOPS, CVPRW 2022}, IEEE Computer Society Conference on
  Computer Vision and Pattern Recognition Workshops, 466--475. IEEE; CVF; IEEE
  Comp Soc.
\newblock ISBN 978-1-6654-8739-9.
\newblock IEEE/CVF Conference on Computer Vision and Pattern Recognition
  (CVPR), New Orleans, LA, JUN 18-24, 2022.

\bibitem[{Sorrentino et~al.(2022)Sorrentino, Sannino, Spinelli, Piana, Boselli,
  Tontodonato, Castellano, and Wang}]{amt-15-149-2022}
Sorrentino, A.; Sannino, A.; Spinelli, N.; Piana, M.; Boselli, A.; Tontodonato,
  V.; Castellano, P.; and Wang, X. 2022.
\newblock A Bayesian parametric approach to the retrieval of the atmospheric
  number size distribution from lidar data.
\newblock \emph{Atmospheric Measurement Techniques}, 15(1): 149--164.

\bibitem[{Sun et~al.(2024)Sun, Ren, Liu, Park, Wang, and
  Cao}]{sun2024ensirensemblealgorithmimage}
Sun, S.; Ren, W.; Liu, Z.; Park, H.; Wang, R.; and Cao, X. 2024.
\newblock EnsIR: An Ensemble Algorithm for Image Restoration via Gaussian
  Mixture Models.
\newblock arXiv:2410.22959.

\bibitem[{Sun et~al.(2022)Sun, Kolbeck, Abshire, Kawa, and
  Mao}]{essd-14-3821-2022}
Sun, X.; Kolbeck, P.~T.; Abshire, J.~B.; Kawa, S.~R.; and Mao, J. 2022.
\newblock Attenuated atmospheric backscatter profiles measured by the CO$_{2}$
  Sounder lidar in the 2017 ASCENDS/ABoVE airborne campaign.
\newblock \emph{Earth System Science Data}, 14(8): 3821--3833.

\bibitem[{Suvorov et~al.(2022)Suvorov, Logacheva, Mashikhin, Remizova, Ashukha,
  Silvestrov, Kong, Goka, Park, and Lempitsky}]{suvorov2022resolution}
Suvorov, R.; Logacheva, E.; Mashikhin, A.; Remizova, A.; Ashukha, A.;
  Silvestrov, A.; Kong, N.; Goka, H.; Park, K.; and Lempitsky, V. 2022.
\newblock Resolution-robust large mask inpainting with fourier convolutions.
\newblock In \emph{Proceedings of the IEEE/CVF winter conference on
  applications of computer vision}, 2149--2159.

\bibitem[{Swales, Pincus, and Bodas-Salcedo(2018)}]{gmd-11-77-2018}
Swales, D.~J.; Pincus, R.; and Bodas-Salcedo, A. 2018.
\newblock The Cloud Feedback Model Intercomparison Project Observational
  Simulator Package: Version 2.
\newblock \emph{Geoscientific Model Development}, 11(1): 77--81.

\bibitem[{Szwarcman et~al.(2025)Szwarcman, Roy, Fraccaro, Þorsteinn
  Elí~Gíslason, Blumenstiel, Ghosal, de~Oliveira, de~Sousa~Almeida, Sedona,
  Kang, Chakraborty, Wang, Gomes, Kumar, Truong, Godwin, Lee, Hsu, Asanjan,
  Mujeci, Shidham, Keenan, Arevalo, Li, Alemohammad, Olofsson, Hain, Kennedy,
  Zadrozny, Bell, Cavallaro, Watson, Maskey, Ramachandran, and
  Moreno}]{szwarcman2025prithvieo20versatilemultitemporalfoundation}
Szwarcman, D.; Roy, S.; Fraccaro, P.; Þorsteinn Elí~Gíslason; Blumenstiel,
  B.; Ghosal, R.; de~Oliveira, P.~H.; de~Sousa~Almeida, J.~L.; Sedona, R.;
  Kang, Y.; Chakraborty, S.; Wang, S.; Gomes, C.; Kumar, A.; Truong, M.;
  Godwin, D.; Lee, H.; Hsu, C.-Y.; Asanjan, A.~A.; Mujeci, B.; Shidham, D.;
  Keenan, T.; Arevalo, P.; Li, W.; Alemohammad, H.; Olofsson, P.; Hain, C.;
  Kennedy, R.; Zadrozny, B.; Bell, D.; Cavallaro, G.; Watson, C.; Maskey, M.;
  Ramachandran, R.; and Moreno, J.~B. 2025.
\newblock Prithvi-EO-2.0: A Versatile Multi-Temporal Foundation Model for Earth
  Observation Applications.
\newblock arXiv:2412.02732.

\bibitem[{Tian et~al.(2017)Tian, Liu, Wang, Li, Yu, and Chu}]{TIAN2017141}
Tian, J.; Liu, J.; Wang, J.; Li, C.; Yu, F.; and Chu, Z. 2017.
\newblock A spatio-temporal evaluation of the WRF physical parameterisations
  for numerical rainfall simulation in semi-humid and semi-arid catchments of
  Northern China.
\newblock \emph{Atmospheric Research}, 191: 141--155.

\bibitem[{Waldmann et~al.(2025)Waldmann, Shah, Wang, Lehmann, Stewart, Xiong,
  Zhu, Bauer, and Chuang}]{waldmann2025panopticonadvancinganysensorfoundation}
Waldmann, L.; Shah, A.; Wang, Y.; Lehmann, N.; Stewart, A.~J.; Xiong, Z.; Zhu,
  X.~X.; Bauer, S.; and Chuang, J. 2025.
\newblock Panopticon: Advancing Any-Sensor Foundation Models for Earth
  Observation.
\newblock arXiv:2503.10845.

\bibitem[{Wang et~al.(2020)Wang, Bentivegna, Zhou, Klein, and
  Elmegreen}]{wang2020physicsinformedneuralnetworksuper}
Wang, C.; Bentivegna, E.; Zhou, W.; Klein, L.; and Elmegreen, B. 2020.
\newblock Physics-Informed Neural Network Super Resolution for
  Advection-Diffusion Models.
\newblock arXiv:2011.02519.

\bibitem[{Wang et~al.(2024)Wang, Gao, Zhao, Sun, and
  Dai}]{wang2024auxiliarylossfreeloadbalancingstrategy}
Wang, L.; Gao, H.; Zhao, C.; Sun, X.; and Dai, D. 2024.
\newblock Auxiliary-Loss-Free Load Balancing Strategy for Mixture-of-Experts.
\newblock arXiv:2408.15664.

\bibitem[{Wu et~al.(2025)Wu, Liu, Wang, Li, and Huang}]{WU2025111223}
Wu, Z.; Liu, W.; Wang, J.; Li, J.; and Huang, D. 2025.
\newblock FrePrompter: Frequency self-prompt for all-in-one image restoration.
\newblock \emph{Pattern Recognition}, 161: 111223.

\bibitem[{Xu et~al.(2023)Xu, Yan, He, Gao, Yang, Wang, Liu, and
  Liu}]{xu2023spatio}
Xu, T.; Yan, K.; He, Y.; Gao, S.; Yang, K.; Wang, J.; Liu, J.; and Liu, Z.
  2023.
\newblock Spatio-temporal variability analysis of vegetation dynamics in china
  from 2000 to 2022 based on leaf area index: A multi-temporal image
  classification perspective.
\newblock \emph{Remote Sensing}, 15(12): 2975.

\bibitem[{Yang, Zhang, and Karniadakis(2020)}]{doi:10.1137/18M1225409}
Yang, L.; Zhang, D.; and Karniadakis, G.~E. 2020.
\newblock Physics-Informed Generative Adversarial Networks for Stochastic
  Differential Equations.
\newblock \emph{SIAM Journal on Scientific Computing}, 42(1): A292--A317.

\bibitem[{Yang et~al.(2023)Yang, Qin, Ning, Xu, Yao, and You}]{10332123}
Yang, T.; Qin, H.; Ning, B.; Xu, Y.; Yao, Y.; and You, Z. 2023.
\newblock Application of t-SNE and PSO-BPNN for Identification of Penetration
  State in Ship GMAW.
\newblock In \emph{2023 IEEE 6th International Conference on Pattern
  Recognition and Artificial Intelligence (PRAI)}, 223--228.

\bibitem[{Yin et~al.(2024)Yin, Meo, Roy, Cher, Wang, Imhoff, Uijlenhoet, and
  Dauwels}]{yin2024precipitationnowcastingusingphysics}
Yin, J.; Meo, C.; Roy, A.; Cher, Z.~B.; Wang, Y.; Imhoff, R.; Uijlenhoet, R.;
  and Dauwels, J. 2024.
\newblock Precipitation Nowcasting Using Physics Informed Discriminator
  Generative Models.
\newblock arXiv:2406.10108.

\bibitem[{Zhang et~al.(2024)Zhang, Gong, He, Zhu, Sun, and
  Zhang}]{zhang2024uirloraachievinguniversalimage}
Zhang, C.; Gong, D.; He, J.; Zhu, Y.; Sun, J.; and Zhang, Y. 2024.
\newblock UIR-LoRA: Achieving Universal Image Restoration through Multiple
  Low-Rank Adaptation.
\newblock arXiv:2409.20197.

\bibitem[{Zhang et~al.(2021)Zhang, Meng, Zeng, Wei, and Shi}]{9440708}
Zhang, L.; Meng, Q.; Zeng, J.; Wei, X.; and Shi, H. 2021.
\newblock Evaluation of Gaofen-3 C-Band SAR for Soil Moisture Retrieval Using
  Different Polarimetric Decomposition Models.
\newblock \emph{IEEE Journal of Selected Topics in Applied Earth Observations
  and Remote Sensing}, 14: 5707--5719.

\bibitem[{Zhang et~al.(2018)Zhang, Isola, Efros, Shechtman, and Wang}]{2018The}
Zhang, R.; Isola, P.; Efros, A.~A.; Shechtman, E.; and Wang, O. 2018.
\newblock The Unreasonable Effectiveness of Deep Features as a Perceptual
  Metric.
\newblock \emph{IEEE}.

\bibitem[{Zhang and Funkhouser(2018)}]{Zhang_2018_CVPR}
Zhang, Y.; and Funkhouser, T. 2018.
\newblock Deep Depth Completion of a Single RGB-D Image.
\newblock In \emph{Proceedings of the IEEE Conference on Computer Vision and
  Pattern Recognition (CVPR)}.

\bibitem[{Zhang et~al.(2023)Zhang, Chen, Jamet, Dionisi, Hu, Lu, and
  Pan}]{2023Retrieving}
Zhang, Z.; Chen, P.; Jamet, C.; Dionisi, D.; Hu, Y.; Lu, X.; and Pan, D. 2023.
\newblock Retrieving bbp and POC from CALIOP: A deep neural network approach.
\newblock \emph{Remote Sensing of Environment}.

\bibitem[{Zhong et~al.(2023)Zhong, Ma, Yao, Xu, Wu, and Wang}]{gmd-16-199-2023}
Zhong, X.; Ma, Z.; Yao, Y.; Xu, L.; Wu, Y.; and Wang, Z. 2023.
\newblock WRF--ML v1.0: a bridge between WRF v4.3 and machine learning
  parameterizations and its application to atmospheric radiative transfer.
\newblock \emph{Geoscientific Model Development}, 16(1): 199--209.

\end{thebibliography}

\end{document}